# A deep learning algorithm for reducing false positives in screening mammography

Stefano Pedemonte[1]*, Trevor Tsue[1], Brent Mombourquette[1], Yen Nhi Truong Vu[1], Thomas Matthews[1], Rodrigo Morales Hoil[1], Meet Shah[1], Nikita Ghare[1], Naomi Zingman-Daniels[1], Susan Holley[2], Catherine M. Appleton[3], Jason Su[1], and Richard L. Wahl[4]

[1] Whiterabbit.ai, Santa Clara, CA, USA
[2] ONsite Women's Health, Westfield, MA, USA
[3] SSM Health, St. Louis, MO, USA
[4] Mallinckrodt Institute of Radiology, Washington University in St. Louis, St. Louis, MO, USA

## Abstract

Screening mammography improves breast cancer outcomes by enabling early detection and treatment. However, false positive callbacks for additional imaging from screening exams cause unnecessary procedures, patient anxiety, and financial burden. This work demonstrates an AI algorithm that reduces false positives by identifying mammograms not suspicious for breast cancer. We trained the algorithm to determine the absence of cancer using 123,248 2D digital mammograms (6,161 cancers) and performed a retrospective study on 14,831 screening exams (1,026 cancers) from 15 US and 3 UK sites. Retrospective evaluation of the algorithm on the largest of the US sites (11,592 mammograms, 101 cancers) a) left the cancer detection rate unaffected (p=0.02, non-inferiority margin 0.25 cancers per 1000 exams), b) reduced callbacks for diagnostic exams by 31.1% compared to standard clinical readings, c) reduced benign needle biopsies by 7.4%, and d) reduced screening exams requiring radiologist interpretation by 41.6% in the simulated clinical workflow. This work lays the foundation for semi-autonomous breast cancer screening systems that could benefit patients and healthcare systems by reducing false positives, unnecessary procedures, patient anxiety, and expenses.

## 1 Introduction

Breast cancer is the most common cancer in women worldwide and the leading cause of cancer deaths in women [1,2]. Regular screening mammograms allow early cancer detection, improve prognosis, and reduce breast cancer mortality [3–7]. Thus, the US and many other nations have developed screening mammography programs [8,9], with the US alone performing over 38 million exams each year [10,11].

While population-based screening mammography has been shown to reduce mortality in participating populations [3–7], sensitivity and specificity vary amongst radiologists [12]. In the US, current mammographic screening is estimated to miss 13% of cancers [12]. Missed diagnoses result in presentations with more advanced disease with higher rates of metastasis, worse disease prognosis, and higher treatment costs [13]. Additionally, false positives constitute a significant concern for patients who undergo breast cancer screening. Over the course of 10 screening exams, more than half of women experience false positive callbacks, and more than 20% are subjected to potentially unnecessary biopsies [11,14]. This results in unnecessary follow-up diagnostic exams, invasive diagnostic procedures, and patient anxiety [15,16]. False positives also constitute a significant expenditure for healthcare systems [17,18], estimated at $2.8 billion annually in the US [18].

Three types of computer aided detection and diagnosis software (CAD) are in use to assist radiologists in detecting breast cancer, with the aim of improving cancer detection performance: 1) software that triages exams by providing an overall malignancy score for each exam (the Food and Drug Administration (FDA) identifies them as CADt, for computer aided triage) [19,20], 2) software that highlights suspicious areas in mammograms by means of bounding boxes, or other image markers (the FDA identifies them as CADe, for computer aided detection) [19,21], and 3) software that provides a second interpretation [19,20,22] in double reading settings, such as in the UK and Europe [23]. Although differing in the interface with radiologists, these three types of software leverage similar technology, a cancer detection and classification algorithm. The third type of software is not regulated in the US and presents a similar mode of operation to CADt. To provide assistance to radiologists, these three software types are designed using an underlying cancer algorithm at an operating point of balanced sensitivity and specificity, similar to how human readers are known to operate. At such a point of balanced sensitivity and specificity, state of the art cancer detection algorithms achieve a performance comparable to average human readers, at approximately 85% sensitivity and specificity [24]. The interaction and cooperation mechanisms between human readers and cancer detection algorithms that operate at balanced sensitivity and specificity, however, leave room for misinterpretation. The low sensitivity (not all malignant findings are highlighted in CADe, and a proportion of exams with cancer are given a low score in CADt) prevents radiologists from focusing solely on highlighted image areas and triaged exams, and the low specificity (many image areas are highlighted in CADe and many negative exams are given high scores in CADt) leads to inefficiency and increased false positives [25]. Consequently, these systems may be prone to causing errors, as the absence of CADe findings or a low CADt score may be incorrectly interpreted as negative assessments and the presence of CADe findings or a high CADt score may be incorrectly interpreted as true positives [25]. The room for misinterpretation of the CAD results may be the reason for the failure of CADe to increase sensitivity of breast cancer screening and decrease false positive rates reliably [12].

Recently, devices that rule out cancer have emerged as a novel paradigm in breast cancer screening [26–28]. Rule-out devices are based on a cancer detection algorithm operated at an extreme operating point at high sensitivity rather than an operating point that balances sensitivity and specificity. The high sensitivity (low false negative rate) operating point enables a more reliable interpretation of the model, where all or nearly all cases marked as non-suspicious are truly cancer-free so that these cases potentially can be removed from the radiologist workflow. This leads to human/machine cooperation with stronger performance guarantees. Rodriquez-Ruiz et al (2019) [26], Yala et al. (2019) [27], Dembrower et al. (2020) [28], Salim et al (2020) [29], and Raya-Povedano et al. (2021) [30] have proposed rule-out devices that can automatically declare 17.0%, 19.3%, 60.0%, 34.1%, and 30.9% of the mammograms as non-suspicious with a sensitivity of 99.0%, 99.0%, 80.1%, 99.0%, and 97.8%, respectively. These works have demonstrated that rule-out devices can automate part of mammographic assessments. In this work, we describe a highly optimized rule-out model that not only enables the automation of a large fraction of exams but also directly benefits patients by reducing false positive callbacks and biopsies. We demonstrate this in an extensive retrospective study based on a diverse dataset of screening exams acquired from large academic hospitals and small outpatient clinics in the US and UK.

# 2 Methods

## 2.1 Retrospective clinical evaluation of the rule-out device

This study uses anonymized data from three institutions to train and evaluate a deep learning system that identifies mammograms that are not suspicious for breast cancer. The objective is to assess how such a cancer rule-out device could impact patients' care and radiologists' workflows. This study was reviewed and approved by the relevant institutional review boards.

## 2.2 Data

Two-dimensional full-field digital mammography (FFDM) exams were gathered from three institutions: two in the US and one in the UK. The US data comes from Washington University in St. Louis (WUSTL) and the multi-site mammography operator ONsite (ONSITE). The UK data comes from the Cancer Research UK OPTIMAM database (OPTIMAM) [31].

The WUSTL and OPTIMAM datasets were randomly divided into three datasets at the patient level without overlap: a training set for training the cancer detection model (80%), a validation set for tuning hyperparameters and selecting operating points (10%), and a test set held out for the evaluation study (10%). The ONSITE dataset was held out for testing. Two scanner models were used to acquire the mammograms: Hologic Selenia (HS) and Hologic Selenia

Dimensions (SD). We leveraged only FFDM images (i.e., no tomosynthesis images). Table 1 summarizes the composition of the datasets. A detailed composition is provided in Supplementary Information S.1.

Diagnostic outcome labels were assigned to exams, single images, and bounding boxes denoting image findings. Exam, image, and bounding box labels were used for model development, while only exam labels were used in the evaluation study. For a complete description of the data labeling methodology, see Supplementary Information S.1. We assigned one of the following labels to each breast: negative (N), corresponding to BI-RADS® 1; screening benign (S), corresponding to BI-RADS® 2; diagnostic benign (D), corresponding to BI-RADS® 0 followed by a negative diagnostic assessment; pathology benign (P), corresponding to a benign pathology assessment; high-risk (H), corresponding to a non-upstaged high-risk pathology assessment; malignant (M), corresponding to a malignant pathology assessment, and interval cancers (I), defined as BI-RADS® 1 or 2 with a malignant pathology assessment within the screening interval (12 months US, 36 months UK) and prior to the subsequent screening exam. Labels not based on pathology assessments (N, S, and D) were additionally required to have a follow-up exam at least two years later and no biopsy events in the entire patient history to ensure exact labels. Breasts not satisfying any of the outlined conditions were labeled unknown (U). The labels were propagated to exam level by selecting the highest priority outcome from the two breasts from N (lowest priority), S, D, U, P, H, I, to M (highest priority). Unknown exams (U) were excluded from all datasets.

| Site / Exam outcome | Washington University in St. Louis, US WUSTL | Multi-site mammography operator, US ONSITE | NHS OPTIMAM, UK OPTIMAM | Model development | Retrospective study, US WUSTL | Retrospective study, US ONSITE | Retrospective study, UK OPTIMAM |
|---|---|---|---|---|---|---|---|
| Negative (N) | 91,413 | 362 | 11,253 | 92,556 | 9,005 | 362 | 1,105 |
| Screening benign (S) | 11,121 | 333 | – | 10,033 (128 annotated) | 1,088 | 333 | – |
| Diagnostic benign (D) | 12,208 | 306 | 295 | 11,264 (2,426 annotated) | 1,217 | 306 | 22 |
| Pathology benign (P) | 1,638 | 31 | 1,525 | 2,855 (910 annotated) | 153 | 31 | 155 |
| Pathology high-risk (H) | 407 | – | – | 379 (304 annotated) | 28 | – | – |
| Malignant (M) | 979 | 330 | 5,334 | 5,663 (4,843 annotated) | 95 | 330 | 555 |
| Interval cancer (I) | 78 | – | 466 | 498 (7 annotated) | 6 | – | 40 |
| Total | 117,844 | 1,362 | 18,873 | 123,248 (8,617 annotated) | 11,592 | 1,362 | 1,877 |

**Table 1**: Data utilized for model development and for the evaluation study. The datasets comprised a total of 138,079 screening exams. Of these, 123,248 from the US and UK were utilized for training and tuning the model, and 14,831 were utilized in the evaluation study that evaluated the effect of the rule-out device on patient care and radiologists' workflow. Findings annotated with bounding boxes, used for model development, are reported in parentheses. A dash (-) indicates information that was not available due to the characteristics of the clinical data records, as specified in Supplementary Information S.1.

## 2.3 Development of the rule-out algorithm

The cancer detection algorithm is composed of two software levels. A low-level vision system based on deep learning models analyzes each image in a mammogram independently. A high-level vision system based on a metamodel combines the information from the low-level models to compute a final malignancy probability for the entire screening exam. Such an architecture enables the algorithm to 1) learn to utilize multi-view, bilateral, and prior imaging data, 2) integrate imaging and non-imaging information, 3) learn from both radiology reports and pathology outcomes obtained through screening, diagnostic, and biopsy events in the clinical workflow. The cancer detection algorithm produces the probability of cancer for a mammogram operating on four inputs: the mammogram images, the patient's age, the prior mammogram's images, and the prior BI-RADS® assessment. Prior images and outcomes were available for 71.2% of WUSTL exams, 35.8% of OPTIMAM exams, and 4.1% of ONSITE exams. Supplementary Information S.2 describes the model architecture and optimization in detail. Supplementary Information S.3 presents an analysis of the contributions of the architecture components. The malignancy prediction algorithm executes in 64.5 seconds per exam using a single GPU NVIDIA GeForce RTX 2080Ti and four modern server CPUs. This run time indicates that the model is suitable for real-time operation within the clinical workflow.

## 2.4 Rule-out workflow and operating point selection

In the simulation of the workflow based on the rule-out device, the device reads the exams before radiologists and assigns a negative score (BI-RADS® 1) to the screening exams with predictions lower than a rule-out operating threshold $T$. For all other exams, the original assessment from the clinical workflow is maintained.

After model training, we calculated the rule-out threshold using the WUSTL and OPTIMAM validation datasets. The threshold was calculated to achieve a 12-month prediction targeted cancer sensitivity, with a method described in detail in Supplementary Information S.4. The evaluation study was performed at two target sensitivity levels: 0.99 and 0.97, resulting in thresholds $T$ = 0.02094 and $T$ = 0.03584. Supplementary Information S.4 characterizes the device performance for a broader range of sensitivity levels between 0.9 and 1.0.

## 2.5 Outcomes

Biopsy-ascertained breast cancer outcomes were utilized as ground truth to measure the radiologist and rule-out device performance. We predicted whether the patient developed cancer within a time window W from the screening exam. We considered prediction windows W = 6, 12, 24 months for the US and W = 6, 12, 24, 36 months for the UK. The cancer-positive class contained malignant, upstaged high-risk, and interval cancer outcomes (i.e., M and I). The negative class contained the exams with outcomes N, S, D, P, and H (Supplementary Information S.7 provides a comprehensive analysis of the algorithm outputs for the exams with non-upstaged high-risk outcomes).

We characterized the radiologists and the rule-out device+radiologists system by their sensitivity and false-positive rate. We characterized the standalone rule-out device by the absolute and relative sensitivity, rule-out rate, reduction of false positive callbacks, and reduction of incorrect biopsies. Diagnostic exams and biopsies performed on patients with negative outcomes are considered incorrect. The radiologists' true positives are defined as cancer cases with BI-RADS® 0. The rule-out device true positives are defined as the cancer cases with predicted scores greater or equal to the rule-out threshold $T$. The rule-out device+radiologists true positives are defined as the cancer cases assessed as BI-RADS® 0 with a device score greater or equal to $T$. The radiologists' sensitivity is the number of radiologists' true positives over the total number of breast cancers. The absolute rule-out device sensitivity is the number of device true positives over the total number of breast cancers. The relative rule-out device sensitivity is the device sensitivity on cases that radiologists can detect and is calculated as the intersection between the device's true positives and the radiologists' true positives over the radiologists' true positives. The rule-out rate is the number of screening exams with scores lower than $T$, over the total number of screening exams. The cancer detection rate (CDR) is the number of true positives, over the total number of mammograms, reported per 1000 mammograms. Rates of reduction of false positive callbacks and biopsies are calculated as the number of exams assessed as negative by the rule-out device (i.e., with a rule-out device score smaller than $T$) that previously led to one or multiple callbacks or biopsies without resulting in a cancer diagnosis, over the total number of exams that led to these procedures without leading to a cancer diagnosis. Supplementary Information S.5 further discusses the interpretation of absolute and relative sensitivity.

To compensate for biases due to the datasets' enrichment, we rebalanced the datasets when computing the values and confidence intervals (CI) for metrics that are dependent on the prevalence of the subclasses N, S, D, P, H, M, and I. These prevalence adjusted metrics are the area under the receiver operating characteristic curve (AUC), CDR, rule-out rate, rates of reduction of callbacks and rates of reduction of biopsies. For sensitivity and CDR metrics, we reported p-values using a non-inferiority z-test for paired proportions [32] with margins of 5% and 0.25 per 1000 exams, respectively. For specificity, rule-out rate, rates of reduction of callbacks, and rates of reduction of biopsies, we computed bootstrap p-values for a one-sided superiority test through inversion of confidence intervals [33]. We present the details of the methodology in Supplementary Information S.6, which also reports raw data to enable the calculation of unadjusted metrics.

# 3 Results

## 3.1 Standalone radiologists and rule-out device cancer detection performance

Figure 1 reports the radiologists' and rule-out device standalone sensitivity and false positive rates. The ROC curves in Figure 1 show that the cancer detection algorithm may be operated as a standalone device at operating points close to the average performance of radiologists in the US [12] and UK [34] (indicated with a black cross). Instead, in this evaluation study, the rule-out device is operated at an extreme operating point on the right side of the ROC curves, with sensitivity nearing 100%.

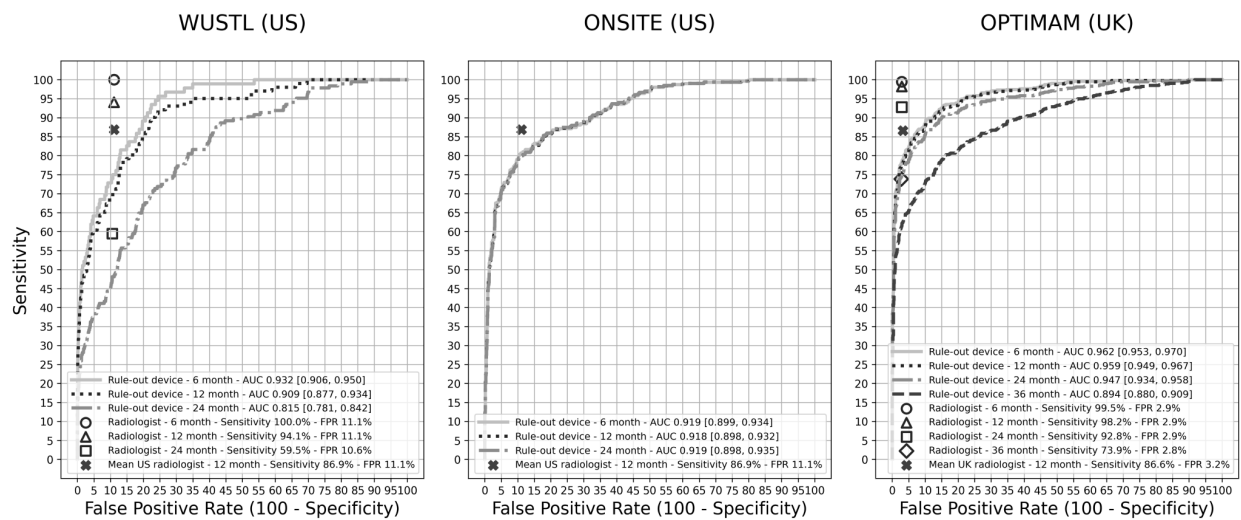

**Figure 1:** Independent device and radiologists sensitivity and false-positive rates calculated for the US and UK datasets to predict cancer over multiple time windows following a screening exam. The black crosses represent average radiologists' performance in the US (BCSC statistics [12]) and the UK (Cancer Research UK statistics [34]). Hollow marks indicate the average radiologists' performance as measured in the evaluation study datasets. Since ONSITE performs primarily screening exams, the radiologists' sensitivity could not be measured as the true extent of false negatives and interval cancers is unknown. The device would perform at a level of performance close to, but not superior to average radiologists if operated as a standalone device at an operating point of balanced sensitivity and specificity. Instead, in this study, the cancer detector is operated as a cancer rule-out device at an extreme operating point on the right side of the ROC curves, with sensitivity nearing 100%.

| | | Target sensitivity 99% | | | | | | | |
|---|---|---|---|---|---|---|---|---|---|
| Scanner | Dataset | Device relative sensitivity (%) | Device absolute sensitivity (%) | Radiologists CDR (per 1000 exams) | Device+Radiologists CDR (per 1000 exams) | Rule-out rate (%) | Decrease false positive callbacks (%) | Decrease benign biopsies (%) | Number of cancers |
| All | US, WUSTL | 100.0 [96.2, 100.0] | 97.0 [91.6, 99.4] | 5.55 [4.49, 6.78] | 5.55 [4.49, 6.78] | 41.6 [40.6, 42.4] | 31.1 [28.7, 33.4] | 7.4 [4.1, 12.4] | 101 |
| All | US, ONSITE | 100.0 [98.9, 100.0] | 100.0 [98.9, 100.0] | 5.76 [5.21, 6.33] | 5.76 [5.21, 6.33] | 19.5 [16.9, 22.1] | 11.9 [8.6, 15.7] | 6.5 [0.0, 19.0] | 330 |
| All | UK, OPTIMAM | 99.6 [98.7, 100.0] | 98.7 [97.4, 99.4] | 9.74 [9.06, 10.44] | 9.71 [9.03, 10.41] | 36.8 [34.4, 39.7] | 17.1 [5.9, 30.1] | 5.9 [2.9, 11.5] | 561 |
| Selenia (HS) | US, WUSTL | 100.0 [76.8, 100.0] | 100.0 [76.8, 100.0] | 5.90 [3.23, 9.89] | 5.90 [3.23, 9.89] | 54.0 [52.4, 55.6] | 42.3 [38.4, 46.6] | 9.5 [2.4, 21.6] | 14 |
| Selenia (HS) | US, ONSITE | 100.0 [94.6, 100.0] | 100.0 [94.7, 100.0] | 5.81 [4.58, 7.23] | 5.81 [4.58, 7.23] | 36.5 [30.9, 41.9] | 26.1 [19.3, 35.2] | 12.5 [0.0, 60.0]* | 68 |
| Selenia (HS) | UK, OPTIMAM | 99.8 [98.9, 100.0] | 98.9 [97.6, 99.6] | 9.75 [9.04, 10.49] | 9.73 [9.02, 10.47] | 37.2 [34.5, 40.1] | 17.9 [7.4, 36.3] | 6.2 [3.2, 11.6] | 519 |
| Dimensions (SD) | US, WUSTL | 100.0 [95.5, 100.0] | 96.6 [90.3, 99.3] | 5.49 [4.37, 6.82] | 5.49 [4.37, 6.82] | 34.9 [33.8, 35.9] | 23.6 [21.2, 26.3] | 6.8 [3.0, 11.9] | 87 |
| Dimensions (SD) | US, ONSITE | 100.0 [98.6, 100.0] | 100.0 [98.6, 100.0] | 5.74 [5.15, 6.37] | 5.74 [5.15, 6.37] | 10.0 [7.7, 12.8] | 4.1 [2.0, 7.7] | 4.3 [0.0, 25.0] | 262 |
| Dimensions (SD) | UK, OPTIMAM | 97.6 [87.4, 99.9] | 95.7 [85.2, 99.5] | 9.59 [7.52, 11.59] | 9.36 [7.30, 11.37] | 13.8 [4.3, 34.8] | 0.0* | 0.0* | 42 |
| | | Target sensitivity 97% | | | | | | | |
| Scanner | Dataset | Device relative sensitivity (%) | Device absolute sensitivity (%) | Radiologists CDR (per 1000 exams) | Device+Radiologists CDR (per 1000 exams) | Rule-out rate (%) | Decrease false positive callbacks (%) | Decrease benign biopsies (%) | Number of cancers |
| All | US, WUSTL | 97.9 [92.6, 99.7] | 95.0 [88.8, 98.4] | 5.55 [4.49, 6.78] | 5.43 [4.39, 6.65] | 56.4 [55.3, 57.2] | 43.5 [41.1, 46.4] | 12.0 [7.4, 17.5] | 101 |
| All | US, ONSITE | 99.4 [97.8, 99.9] | 99.4 [97.8, 99.9] | 5.76 [5.21, 6.33] | 5.72 [5.18, 6.29] | 31.6 [28.4, 34.6] | 22.7 [18.4, 27.3] | 12.9 [3.3, 27.8] | 330 |
| All | UK, OPTIMAM | 98.9 [97.6, 99.6] | 97.0 [95.3, 98.2] | 9.74 [9.06, 10.44] | 9.64 [8.96, 10.34] | 51.7 [49.1, 54.6] | 29.6 [14.6, 43.5] | 7.2 [4.1, 12.5] | 561 |
| Selenia (HS) | US, WUSTL | 100.0 [76.8, 100.0] | 100.0 [76.8, 100.0] | 5.90 [3.23, 9.89] | 5.90 [3.23, 9.89] | 67.9 [66.4, 69.3] | 54.6 [51.0, 59.2] | 9.5 [2.4, 21.6] | 14 |
| Selenia (HS) | US, ONSITE | 97.0 [89.6, 99.6] | 97.1 [89.8, 99.6] | 5.81 [4.58, 7.23] | 5.64 [4.43, 7.04] | 48.3 [43.1, 54.5] | 43.6 [34.3, 52.4] | 37.5 [0.0, 75.0]* | 68 |
| Selenia (HS) | UK, OPTIMAM | 99.2 [98.0, 99.8] | 97.4 [95.8, 98.6] | 9.75 [9.04, 10.49] | 9.68 [8.97, 10.41] | 52.2 [49.2, 55.0] | 31.1 [18.9, 47.5] | 7.6 [3.7, 13.2] | 519 |
| Dimensions (SD) | US, WUSTL | 97.5 [91.4, 99.7] | 94.3 [87.1, 98.1] | 5.49 [4.37, 6.82] | 5.36 [4.25, 6.67] | 50.1 [48.9, 51.2] | 35.9 [32.4, 38.7] | 12.8 [7.5, 19.4] | 87 |
| Dimensions (SD) | US, ONSITE | 100.0 [98.6, 100.0] | 100.0 [98.6, 100.0] | 5.74 [5.15, 6.37] | 5.74 [5.15, 6.37] | 22.3 [19.3, 26.7] | 11.6 [8.0, 16.8] | 4.3 [0.0, 25.0] | 262 |
| Dimensions (SD) | UK, OPTIMAM | 95.2 [83.8, 99.4] | 91.3 [79.2, 97.6] | 9.59 [7.52, 11.59] | 9.13 [7.08, 11.16] | 27.5 [10.8, 48.1] | 0.0* | 0.0* | 42 |

**Table 2:** Standalone rule-out device, radiologists, and rule-out device+radiologists cancer detection performance metrics for WUSTL, ONSITE, and OPTIMAM test datasets. The prediction windows are 12 months in the US and 24 months in the UK. Section

2.5 Outcomes defines the metrics, and Supplementary Information S.5 defines the statistical methods for calculating the 95% confidence intervals. (*) indicates less than ten samples.

## 3.2 Effect of the rule-out device on quality of screening for patients

Table 2 summarizes the main results of this study for target sensitivities of 99% and 97%. Results are reported for the 12-month prediction window for the US and 24-month for the UK. We focus primarily on the 99% sensitivity operating point. At this operating point, the rule-out device had a relative sensitivity of 100% [95% CI 96.2%, 100%] in WUSTL, 100% [98.9%, 100%] in ONSITE, and 99.6% [98.7%, 100%] in OPTIMAM. Two cancers were missed across the three datasets that would have been detected without the rule-out device (images and retrospective reviews of false negative exams are included in Supplementary Information S.10). The sensitivity of the screening workflow based on the rule-out device (not reported in the table but can be calculated as the product of the sensitivity of the standard workflow and the relative sensitivity of the device) was not inferior to the sensitivity of the standard workflow (WUSTL: $p = 0.01$; OPTIMAM: $p < 0.001$; ONSITE: $p < 0.001$ with 5% non-inferiority margin). As a consequence of the high sensitivity, the cancer detection rate (CDR) was unaffected by the rule-out device, within the non-inferiority margin of 0.25 detections per 1000 exams, in all three datasets (WUSTL: $p = 0.02$; ONSITE: $p < 0.001$; OPTIMAM: $p < 0.001$). The rule-out device marked the following percentage of mammograms as non-suspicious: 41.6% [40.6%, 42.4%] in WUSTL, 19.5% [16.9%, 22.1%] in ONSITE and 36.8% [34.4%, 39.7%] in OPTIMAM. The device marked as non-suspicious several mammograms that in the clinical workflow were false positives. By marking these cases as non-suspicious, the device reduced the false positive callbacks by 31.1% [28.7%, 33.4%] in WUSTL, 11.9% [8.6%, 15.7%] in ONSITE, and 17.1% [5.9%, 30.1%] in OPTIMAM. These reduction rates were significantly larger than 0 ($p<0.001$ WUSTL, $p<0.001$ ONSITE, $p<0.001$ OPTIMAM). Similarly, the rule-out device marked as negative several screening exams that in the clinical workflow had led to biopsies with negative outcomes. By marking these mammograms as non-suspicious, the rule-out device reduced the number of negative biopsies by 7.4% [4.1%, 12.4%] in WUSTL, 6.5% [0.0%, 19.0%] in ONSITE, and 5.9% [2.9%, 11.5%] in OPTIMAM. These reduction rates were significantly larger than 0 for WUSTL and OPTIMAM ($p<0.001$ WUSTL, $p<0.001$ OPTIMAM; $p=0.082$ for ONSITE). Table 2 and the plots in Figure 2 report these results in further detail and by individual scanner model.

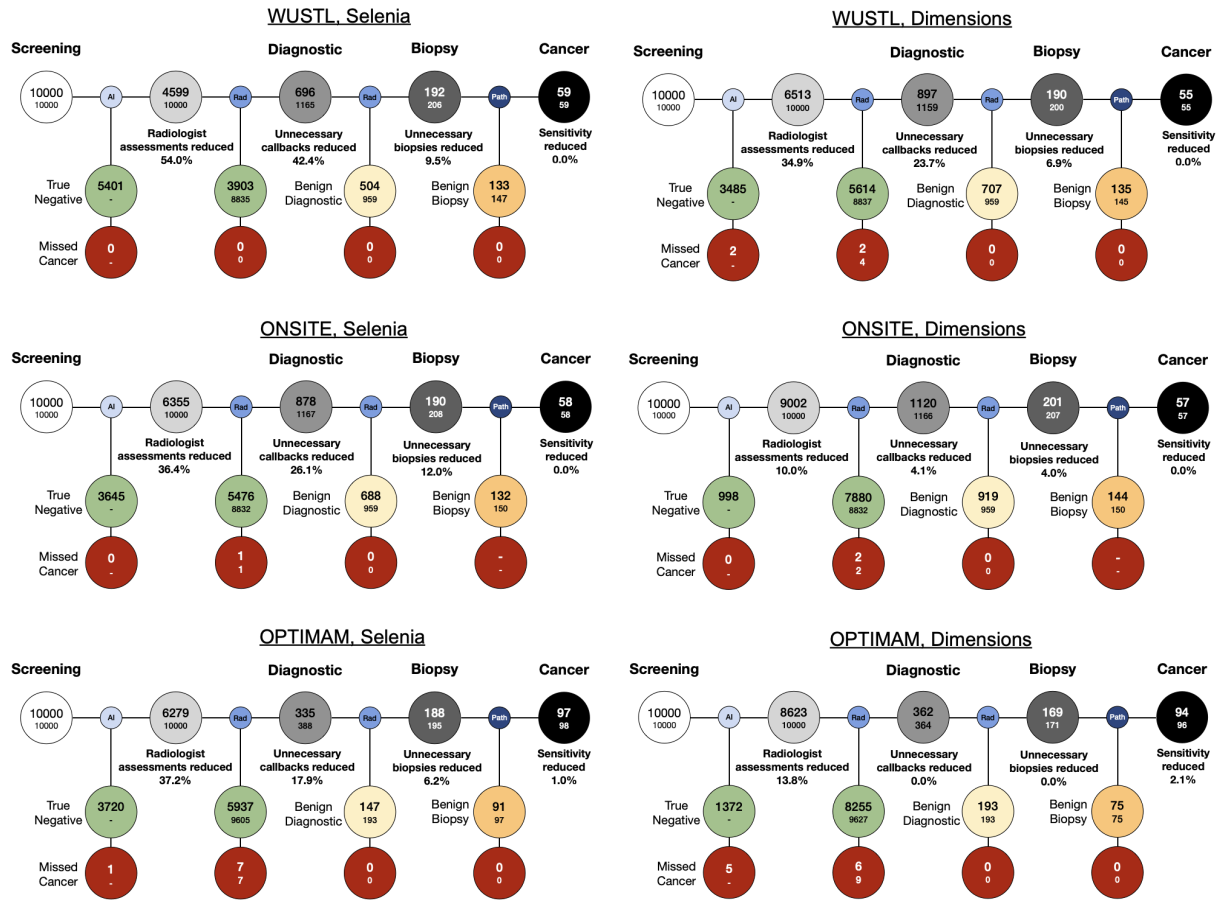

**Figure 2**: Effect of the rule-out device on the screening workflow. Workflow normalized to 10,000 screening exams. The rule-out device is operated at a target sensitivity of 99%, and the outcomes are based on information recorded within 12 months from the screening exams for the US and 24 months for the UK. In each node, the small font represents the standalone radiologists, and the large font the rule-out device+radiologists workflow. By marking a subset of the screening exams as non-suspicious, the rule-out device has the downstream effect of reducing false positive callbacks (reduction greater than 0; WUSTL: p<0.001; ONSITE: p<0.001; OPTIMAM: p<0.001) and biopsies (reduction greater than 0; WUSTL: p<0.001; OPTIMAM: p<0.001 OPTIMAM; ONSITE: not significant p=0.082) while maintaining sensitivity (WUSTL: p=0.01; OPTIMAM: p<0.001; ONSITE: p<0.001 with 5% non-inferiority margin). The percent reduction metrics differ slightly (within the first decimal) from the metrics reported in Table 2 due to integer rounding errors in the normalization. Supplementary Information, S.6 reports un-normalized values.

The differences in per-scanner performance are likely attributable to the differences in the number of cancer-positive mammograms in the development dataset: 4,867 for scanner model HS and 1,294 for SD. Detailed data breakdown of the model performance by scanner model is presented in Supplementary Information S.1. Examples of patient callbacks and biopsies prevented by the rule-out system are presented in Supplementary Information S.11.

Figure 3 presents the device's standalone sensitivity and the effect on the collective radiologists' sensitivity for the prediction of cancer over multiple time windows. These values are calculated using data from both scanner models. Numerical values and 95% confidence intervals for the sensitivity metrics graphed in Figure 3 are reported in Table 2 for the reference prediction windows (12-month for the US and 24-month for the UK). Figure 3 highlights that the loss of sensitivity introduced by the rule-out device, expressed by the difference between the radiologists (diamond markers) and rule-out device+radiologists (large circular markers) sensitivity curves, is slight for all time windows considered, when compared with the inter-reader variability. The inter-reader variability can be inferred from the distribution of the individual radiologists' sensitivity (small circular markers). The absolute rule-out device sensitivity (triangular markers) is lower for more extended time windows due to interval cancers. The rule-out device relative sensitivity (cross markers) remains constant over multiple time windows, confirming that the reduction of absolute sensitivity is due to interval cancers that may not have been visible at the time of screening.

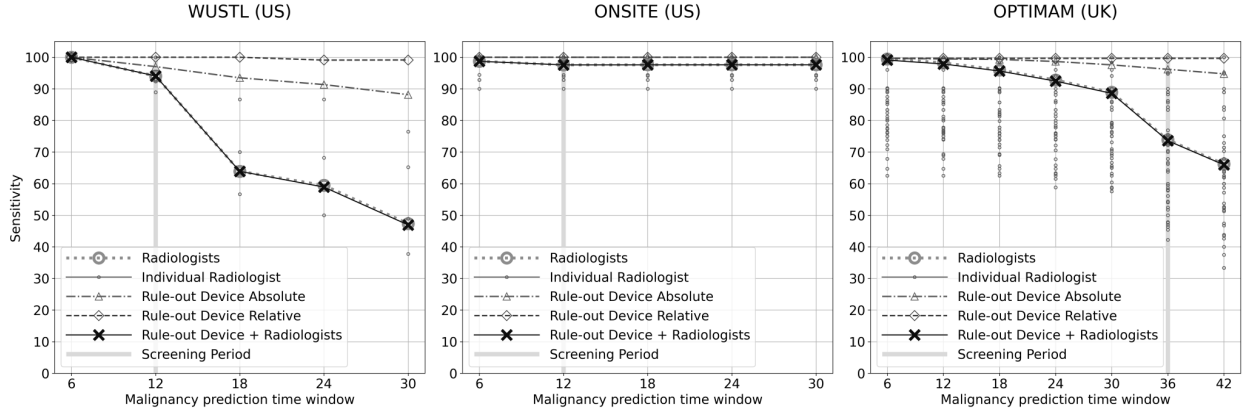

**Figure 3:** Breast cancer sensitivity metrics calculated for cancer prediction over multiple time windows following the screening exams, for the operating point at 99% sensitivity. We report the radiologists' collective absolute sensitivity without the use of the rule-out device (cross) and with the use of the rule-out device (large circle). The small circles represent the sensitivity of individual radiologists. We also report the rule-out device standalone relative (diamond) and absolute (triangle) sensitivity. Numerical values and confidence intervals for each sensitivity metric are reported in Table 2 for the reference windows (12-month for the US and the 24-month for the UK). Since ONSITE performs primarily screening exams, the radiologists' sensitivity could not be measured as the true extent of false negatives and interval cancers is unknown. Radiologists were included in this analysis if they read at least 10 cancer positive screening exams within this study dataset.

## 3.3 Effect of the rule-out device on radiologists' performance

We evaluated the effect of the rule-out device on the individual radiologists who originally read the mammograms and on their collective performance. This analysis included both scanner models. Figure 4 compares individual and collective radiologists' sensitivity and false-positive rates with the use of the rule-out device (diamond marker) and without (circular marker) for the reference windows (12-month for the US and 24-month for the UK). The collective radiologist performance, calculated by considering all readers as one, is displayed with larger markers. A region of acceptable performance, as defined by Lehman et al. [12], is represented as a gray rectangle for the US sites. For the double reading system in OPTIMAM, the rule-out device improved specificity from 94.7% to 96.0% [94.6%, 96.7%] for the first reader and from 97.1% to 97.6% [97.2%, 98.0%] for the last reader, while reducing sensitivity from 82.6% [79.1%, 85.8%] to 82.4% [78.9%, 85.6%] for the first reader and from 92.8% [90.0%, 94.6%] to 92.4% [89.6%, 94.3%] for the last reader. For the single reader system in WUSTL, the rule-out device had a more pronounced effect. The average WUSTL radiologist's specificity increased (p < 0.001) from 88.9% to 92.4% [92.1%, 92.7%] while maintaining sensitivity (p=0.01 with 5% non-inferiority margin) at 94.1% [87.5%, 97.8%]. For ONSITE, the rule-out device increased specificity (p<0.001) from 88.8% to 90.2% [89.8%, 90.6%] and maintained sensitivity at 97.6% [95.3%, 99.0%] (p<0.001 with 5% non-inferiority margin).

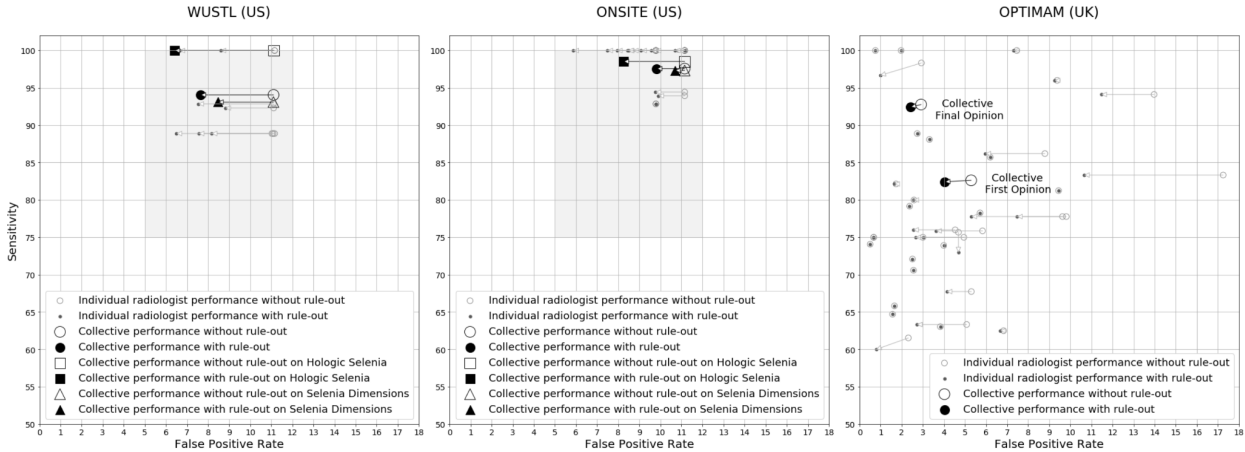

**Figure 4:** Small circular marks report the effect of the rule-out device on the cancer detection performance of individual radiologists. Large circular marks report the effect on their collective performance. In the WUSTL and ONSITE results, large square and triangular marks report the effect of the rule-out device on radiologists' collective performance when reading exams from the Hologic Selenia and Hologic Selenia Dimensions scanners respectively. This breakdown is not reported for the OPTIMAM (UK) dataset, as it consists primarily (95.9%) of exams acquired on Hologic Selenia. In the UK paradigm with multiple readers, the collective first opinion refers to the decision for the first reader, while the collective final opinion refers to the final decision based on radiologists' consensus. The performance of an individual radiologist is based on all opinions provided by the radiologist, as first, second, or arbitrating reader. In the US results, the gray rectangle represents the region of acceptable performance, as defined in Lehman et al. [12]. The UK National Health System (NHS) considers acceptable false positive rates between 3% and 9% and does not define explicit thresholds of acceptable performance for sensitivity, focusing instead on performance thresholds for the cancer detection rate [35]. Outcomes are derived from exams within 12 months from the screening exam for the US and within 24 months for the UK. Radiologists were included in this analysis if they read at least 10 cancer positive and 10 cancer negative screening exams within this study dataset.

# 4 Discussion

The risks of breast cancer screening can be understood in light of the recent change of breast cancer screening recommendations from the US Preventive Services Task Force (USPSTF). To reduce the potential harms associated with false positive mammograms, in 2009 the USPSTF recommended biennial screening starting at age 50, in place of the previous recommendations to screen annually starting at age 40. Implementation of these recommendations was estimated to result in a 68% reduction of false positives at the cost of a 30% reduction of sensitivity and of the number of deaths averted [36]. Consequently, many women will die from breast cancers that would otherwise have been found by more frequent screening. In this study, we have shown that algorithms optimized to identify negative cases may achieve at least half of the reduction of the false positive rate achieved by the USPSTF guidelines update, while introducing an effect on the sensitivity of the order of 1%.

The elimination of incorrect follow-up exams and biopsies, which constitute major limitations of breast cancer screening today, benefits patients directly and is the most critical advantage of cancer rule-out technology. Additionally, with such technology, a large fraction of patients with normal mammograms would be able to immediately receive the cancer-negative report, which could result in increased patient satisfaction and decreased anxiety. A third potential benefit of cancer rule-out technology is to reduce costs and deliver screening results sustainably by enabling the automation of a large fraction of radiological assessments. This can benefit mature screening systems affected by workforce shortages and expand nascent and under-resourced screening systems. Reduced false positives and patient anxiety may also lead to an increase in screening compliance, as false positives and anxiety are linked to lower screening compliance rates [17].

In this evaluation study, we assumed that the radiologist's operating point is unaffected by the removal of exams from their worklist. Reader studies are required to investigate this hypothesis. We anticipate two primary effects: 1) increased prevalence of cancer in the subset of cases assessed by radiologists; 2) knowledge that the rule-out device has eliminated cases that it considers cancer-negative. In practice, due to the rarity of cancer, prevalence will increase only slightly (from a US average of 0.5% in the screening population to approximately 1.0% when we consider a rule-out system that assesses as negative 50% of the negative exams). Therefore, we hypothesize that increased prevalence may not be noticeable and may therefore have a negligible effect on radiologists' interpretation. The second effect may be expected to lead to increased sensitivity, as radiologists will be able to focus on the most challenging cases, although further studies are required to verify this hypothesis. Further studies may also enable the optimization of the collaborative interfaces for rule-out devices in the clinical settings. An alternative to the semi-autonomous hard rule-out system investigated in the present work, which eliminates exams from the radiologist's worklist, is an assistive soft rule-out system that marks negative cases in the worklist, leaving the final decision to radiologists. This latter AI+human approach would likely be a more realistic initial deployment strategy for this system before full autonomy.

The results presented in this study indicate that rule-out devices are potentially safer than existing assistive technologies. If the AI decisions are entirely accepted, as assumed in this study, the system introduces the largest advantages in reducing false positives, callbacks, and biopsies. In contrast, existing CADt and CADe devices operate at a level of sensitivity near 85%; thus, a negative algorithmic assessment cannot be accepted as a true negative. As a result, the effect of existing CADe and CADt devices on sensitivity is highly dependent on the user's perception of the device's characteristics and, therefore, subject to unpredictability.

There are limitations to this study. The cancer cases on Selenia Dimensions scanners were limited in training and testing (except for the ONSITE dataset for testing) and were about ¼ of the number of cancer cases on Hologic Selenia. This imbalance is due to the later introduction of Selenia Dimensions in the market and may have contributed to the lower performance of the algorithm for this scanner model. Outcome information for the ONSITE data was limited. ONsite predominantly performs screening mammography, which makes tracking interval cancers more challenging; this results in an overestimation of radiologists' and absolute device sensitivity for this site. We discuss this further in Supplementary Information, S.5. In estimating sensitivity, radiologists' false negatives were defined based on interval cancers without discerning between missed cancers and true interval cancers that were not present or visible at the time of the screening exam. Additionally, radiologists' false negative cases may be present that were not tracked by the clinics where the screening exams were performed. These two limitations are common to the large majority of studies in mammography as they are due to systemic limitations in breast cancer tracking, especially in the US.

The results presented in this paper, in comparison to the effect of the updated USPSTF guidelines, urge us to consider the introduction of cancer rule-out devices in clinical practice. Quality assurance and monitoring systems must be devised to guarantee a safe operation and further retrospective and prospective investigations are required to substantiate the benefits to patients, radiologists, and the healthcare system. With these in place, rule-out devices could offer a significantly safer and more effective alternative to improving screening than restrictive nationwide guideline changes.

## Competing Interests

This work was partially supported by funding from Whiterabbit AI, Inc. Washington University has equity interests in Whiterabbit AI, Inc. and may receive royalty income and milestone payments from a "Collaboration and License Agreement" with Whiterabbit AI, Inc. to develop a technology evaluated in this research. RLW is the principal investigator on a research contract from White Rabbit AI. These agreements are managed by the WU Institutional COI Committee. In addition, the following authors are employed by and/or have equity interests in Whiterabbit AI, Inc.: S.P., T.T., B.M., Y.N.T.V., T.M., R.M.H., M.S., N.G., N.Z.D., C.M.A., and J.S.

## Acknowledgments

The authors thank Dr. Mark Halling-Brown for elucidating the many aspects of data collection of the OPTIMAM database, Dr. Daniel Marcus, Jenny Gurney, David Maffit, and Stephen Moore for the coordination efforts for data acquisition, and Dominique Ward, Chip Schweiss, and Nate Dolley for the technical assistance.